\documentclass{llncs}
\usepackage{algorithm}
\usepackage{algorithmic}
\usepackage{graphicx}
\begin{document}
\title{Content Enhanced BERT-based \newline Text-to-SQL Generation}
%
%
\author{
Tong Guo\inst{1}
Huilin Gao\inst{2}
}
%

%
\institute{Never Stop Research\and
China Electronic Technology Group Corporation Information Science Academy, Beijing, China}
\maketitle              
\begin{abstract}
We present a simple methods to leverage the table content for the BERT-based model to solve the text-to-SQL problem. Based on the observation that some of the table content match some words in question string and some of the table header also match some words in question string, we encode two addition feature vector for the deep model. Our methods also benefit the model inference in testing time as the tables are almost the same in training and testing time. We test our model on the WikiSQL dataset and outperform the BERT-based baseline by 3.7\% in logic form and 3.7\% in execution accuracy and  achieve state-of-the-art.

\end{abstract}
\section{Introduction}

Semantic parsing is the tasks of translating natural language to logic form. Mapping from natural language to SQL (NL2SQL) is an important semantic parsing
task for question answering system. In recent years, deep learning and BERT-based model have shown significant improvement on this task. However, past methods did not encode the table content for the input of deep model. For industry application, the table of training time and the table of testing time are the same. So the table content can be encoded as external knowledge for the deep model.

In order to solve the problem that the table content is not used for model, we propose our effective encoding methods, which could incorporate database designing information into the model. Our key contribution are three folds:

1. We use the match info of all the table cells and question string to mark the question and produce a feature vector which is the same length to the question. 

2. We use the match inf of all the table column name and question string to mark the column and produce a feature vector which is the same length to the table header. 

3. We design the whole BERT-based model and take the two feature vector above as external inputs. The experiment results outperform the baseline\cite{ref_proc3}. The code is available.\footnote[1]{\url{https://github.com/guotong1988/NL2SQL-RULE}}

\begin{figure}
\centering
\includegraphics[width=\textwidth]{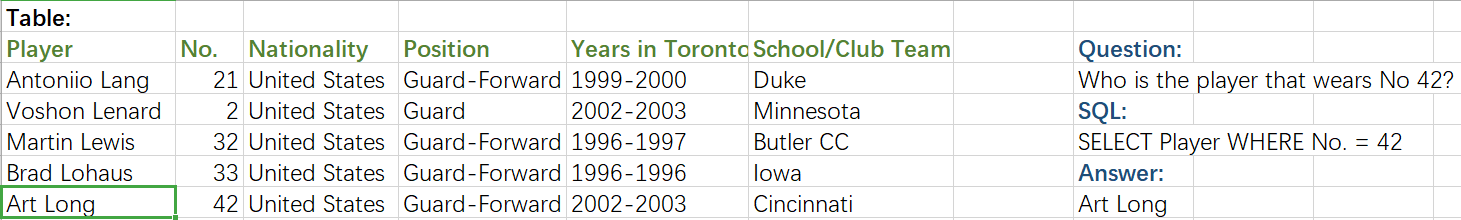}
\caption{An example of WikiSQL dataset} \label{fig1}
\end{figure}

\section{Related Work}

WikiSQL \cite{ref_proc1} is a large semantic parsing dataset. It has 80654 natural language and corresponding SQL pairs. The examples of WikiSQL are shown in fig. 1. 

BERT\cite{ref_proc4} is a very deep transformer-based\cite{ref_proc5} model. It first pre-train on very large corpus using the mask language model loss and the next-sentence loss. And then we could fine-tune BERT on a variety of specific tasks like text classification, text matching and natural language inference and set new state-of-the-art performance on them.

\section{External Feature Vector Encoding}

\begin{algorithm}
\caption{The construction for question mark vector}
\label{alg:A}
\begin{algorithmic}
\STATE{vector = [0]*question\_length}
\FOR{cell in table}
\IF{contains(question,cell)}
\STATE{start\_index = get\_index(question, cell)}
\STATE{vector[start\_index:start\_index+len(cell)] = 2}
\STATE{vector[start\_index] = 1}
\STATE{vector[start\_index+len(cell)] = 3}
\STATE{break}
\ENDIF
\ENDFOR

\FOR{one in header}
\IF{contains(question,one)}
\STATE{index = get\_index(question, one)}
\STATE{vector[index] = 4}
\ENDIF
\ENDFOR

\end{algorithmic}
\end{algorithm}

\begin{algorithm}
\caption{The construction for table header mark vector}
\label{alg:A}
\begin{algorithmic}
\STATE{vector = [0]*header\_length}
\STATE{index = 0}
\FOR{one in header}
\IF{contains(question, one)}
\STATE{vector[index] = 1}
\ENDIF
\STATE{index = index + 1}
\ENDFOR

\FOR{cell in table}
\IF{contains(question,cell)}
\STATE{vector[get\_column\_index(table, cell)] = 2}
\ENDIF
\ENDFOR

\end{algorithmic}
\end{algorithm}

In this section we describe our encoding methods based on the word matching of table content and question string and the word matching of table header and question string. The full algorithms are shown in Algorithm 1 and Algorithm 2. In the Algorithm 1, the value 1 stand for 'START' tag, value 2 stand for 'MIDDLE' tag, value 3 stand for 'END' tag. In the Algorithm 2, we think that the column, which contains the matched cell, should be marked. The final question mark vector is named $QV$ and the final table header mark vector is named $HV$. For industry application, we could refer to Algorithm 1 and Algorithm 2 to encode external knowledge flexibly.

\begin{figure}
\centering
\includegraphics[width=\textwidth]{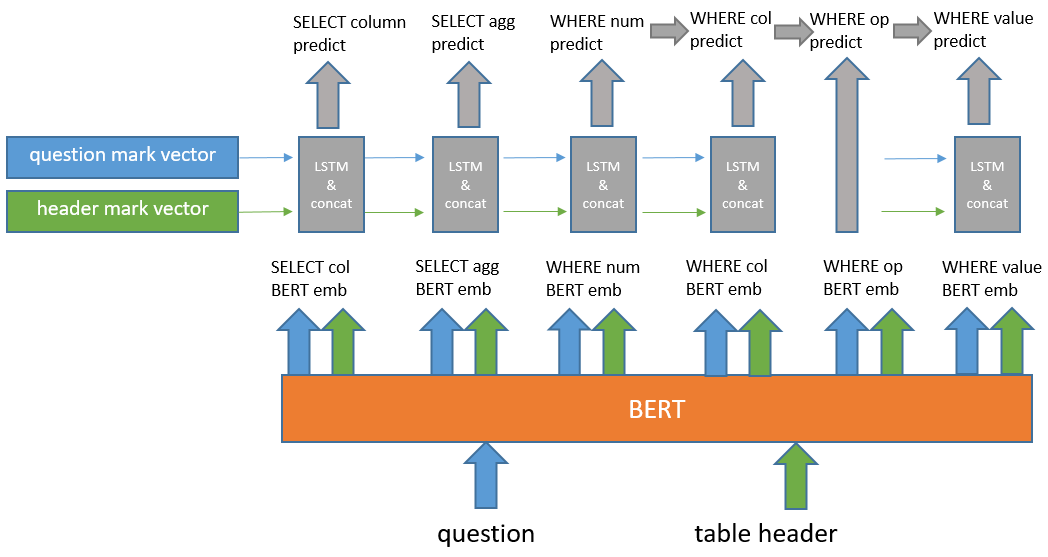}
\caption{The deep model} \label{fig1}
\end{figure}

\section{The Deep Neural Model}

Based on the Wikisql dataset, we also use three sub-model to predict the SELECT part, AGG part and WHERE part. The whole model is shown in fig. 2.

We use BERT as the representation layer. The question and table header are concat and then input to BERT, so that the question and table header have the attention interaction information of each other.
We denote the BERT output of question and table header as $Q$ and $H$

\subsection{BERT embedding layer}

Given the question tokens ${w_1,w_2,...,w_n}$ and the table header ${h_1,h_2,...,h_n}$, we follow the BERT convention and concat the question tokens and table header for BERT input. The detail encoding is below:

$[CLS],w_1,w_2,...,w_n, [SEP], h_1, [SEP], h_2, [SEP],..., h_n,[SEP] $

The output embeddings of BERT are shared in all the downstream tasks. We think the concatenation input for BERT can produce some kind of 'global' attention for the downstream tasks.

\subsection{SELECT column}
Our goal is to predict the column name in the table header. The inputs are the question $Q$ and table header $H$. The output are the probability of SELECT column: 
\begin{equation}
P(sc|Q,H,QV,HV)
\end{equation}
where $QV$ and $HV$ is the external feature vectors that are described above.
\subsection{SELECT agg}
Our goal is to predict the agg slot. The inputs are $Q$  with $QV$ and the output are the probability of SELECT agg:
\begin{equation}
P(sa|Q,QV) 
\end{equation}

\subsection{WHERE number}

Our goal is to predict the where number slot. The inputs are $Q$ and $H$ with $QV$ and $HV$. The output are the probability of WHERE number:
\begin{equation}
P(wn|Q,H,QV,HV) 
\end{equation}

\subsection{WHERE column}

Our goal is to predict the where column slot for each condition of WHERE clause. The inputs are $Q$, $H$ and $P_{wn}$ with $QV$ and $HV$. The output are the top $wherenumber$ probability of WHERE column:
\begin{equation}
P(wc|Q,H,P_{wn},QV,HV) 
\end{equation}

\subsection{WHERE op}

Our goal is to predict the where column slot for each condition of WHERE clause. The inputs are $Q$, $H$, $P_{wc}$ and $P_{wn}$. The output are the probability of WHERE op slot:
\begin{equation}
P(wo|Q,H,P_{wn},P_{wc}) 
\end{equation}

\subsection{WHERE value}
Our goal is to predict the where column slot for each condition of WHERE clause. The inputs are $Q$, $H$, $P_{wn}$, $P_{wc}$ and $P_{wo}$ with $QV$ and $HV$. The output are the probability of WHERE value slot:
\begin{equation}
P(wv|Q,H,P_{wn},P_{wc},P_{wo},QV,HV) 
\end{equation}

\section{Experiments}

In this section we describe detail of experiment parameters and show the experiment result.

\subsection{Experiment result}

In this section, we evaluate our methods versus other approachs on the WikiSQL dataset. See Table 1 and Table 2 for detail. The SQLova\cite{ref_proc3} result use the BERT-Base-Uncased pretrained model and run on our machine without execution-guided decoding(EG)\cite{ref_proc6}.

\begin{table}
\caption{Overall result on the WikiSQL task}\label{tab1}
\centering
\begin{tabular}{|l|l|l|l|l|}
\hline
Model & Logic Form Dev Acc & Execution Dev Acc & Logic Form Test Acc & Execution Test Acc\\
\hline
SQLova\cite{ref_proc3} & 80.6$\%$ & 86.5$\%$ & 80.0$\%$ & 85.5$\%$ \\ 
\hline
Our methods & 84.3$\%$ & 90.3$\%$ & 83.7$\%$ & 89.2$\%$ \\ 
\hline
Our methods + EG & 85.4$\%$ & 91.1$\%$ & 84.5$\%$ & 90.1$\%$ \\ 
\hline
\end{tabular}
\end{table}

\begin{table}
\caption{Break down result on the WikiSQL dataset.}\label{tab1}
\centering
\begin{tabular}{|l|l|l|l|l|l|l|}
\hline
Model & SELECT column & SELECT agg & WHERE number & WHERE column & WHERE op & WHERE value\\
\hline
SQLova\cite{ref_proc3} & 97.0$\%$ & 90.1$\%$ & 98.5$\%$ & 94.4$\%$& 97.3$\%$ & 95.5$\%$ \\ 
\hline
Our methods & 97.4$\%$ & 90.0$\%$ & 99.1$\%$ & 97.9$\%$ & 98.1$\%$ &97.6$\%$\\
\hline
Our methods + EG & 97.4$\%$ & 90.4$\%$ & 98.9$\%$ & 97.9$\%$ & 97.7$\%$ &97.9$\%$\\
\hline
\end{tabular}
\end{table}

\section{Ablation study}
The detail results are shown at Table 3. The header vector mainly improve the result of WHERE OP and the question vector mainly improve the result of WHERE VALUE.

\begin{table}
\caption{The results of ablation study}\label{tab1}
\centering
\begin{tabular}{|l|l|l|l|l|l|l|}
\hline
Model & SELECT col & SELECT agg & WHERE num & WHERE col & WHERE op & WHERE value\\
\hline
without all vector & 97.0$\%$ & 90.1$\%$ & 98.5$\%$ & 94.4$\%$& 97.3$\%$ & 95.5$\%$ \\ 
\hline
without header vector & 97.0$\%$ & 90.1$\%$ & 98.5$\%$ & 94.4$\%$ & 97.3$\%$ &97.3$\%$\\
\hline
without question vector & 97.0$\%$ & 90.1$\%$ & 98.5$\%$ & 97.8$\%$ & 97.3$\%$ &95.5$\%$\\
\hline
all  & 97.4$\%$ & 90.0$\%$ & 99.1$\%$ & 97.9$\%$ & 98.1$\%$ &97.6$\%$\\
\hline
\end{tabular}
\end{table}

\section{Conclusion}

Based on the observation that the table data is almost the same in training time and testing time and to solve the problem that the table content is lack for deep model. We propose a simple encoding methods that can leverage the table content as external feature for the BERT-based deep model, demonstrate its good performance on the WikiSQL task, and achieve state-of-the-art on the datasets.

\end{document}